\documentclass{article}

\usepackage{ijcai22}              

\usepackage[T1]{fontenc}
\usepackage{amsmath,amssymb,amsfonts,mathrsfs,bm}
\usepackage{mathtools}
\usepackage{amsthm}
\usepackage{nicefrac}
\usepackage[shortlabels]{enumitem}
\usepackage{graphicx}
\usepackage{epstopdf}
\usepackage{url}
\usepackage{colortbl}
\usepackage{booktabs}
\usepackage{multirow}
\usepackage[table,dvipsnames]{xcolor}
\usepackage[normalem]{ulem}
\usepackage{xparse}

\usepackage{array}
\newcolumntype{L}[1]{>{\raggedright\let\newline\\\arraybackslash\hspace{0pt}}m{#1}}
\newcolumntype{C}[1]{>{\centering\let\newline\\\arraybackslash\hspace{0pt}}m{#1}}
\newcolumntype{R}[1]{>{\raggedleft\let\newline\\\arraybackslash\hspace{0pt}}m{#1}}

\makeatletter
\let\MYcaption\@makecaption
\makeatother
\usepackage[font=footnotesize]{subcaption}
\makeatletter
\let\@makecaption\MYcaption
\makeatother



\usepackage{glossaries}

\makeatletter
\let\oldgls\gls
\let\oldglspl\glspl

\newcommand\fussy@ifnextchar[3]{%
  \let\reserved@d=#1%
  \def\reserved@a{#2}%
  \def\reserved@b{#3}%
  \futurelet\@let@token\fussy@ifnch}
\def\fussy@ifnch{%
  \ifx\@let@token\reserved@d
    \let\reserved@c\reserved@a 
  \else
    \let\reserved@c\reserved@b
  \fi
 \reserved@c}

\renewcommand{\gls}[1]{%
  \oldgls{#1}\fussy@ifnextchar.{\@checkperiod}{\@}}
\renewcommand{\glspl}[1]{%
  \oldglspl{#1}\fussy@ifnextchar.{\@checkperiod}{\@}}

\newcommand{\@checkperiod}[1]{%
  \ifnum\sfcode`\.=\spacefactor\else#1\fi
}
\makeatother

\newacronym{wrt}{w.r.t.}{with respect to}
\newacronym{RHS}{R.H.S.}{right-hand side}
\newacronym{LHS}{L.H.S.}{left-hand side}
\newacronym{iid}{i.i.d.}{independent and identically distributed}

\usepackage{float}
\ifx\notloadhyperref\undefined
	\ifx\loadbibentry\undefined
		\usepackage[hidelinks,hypertexnames=false]{hyperref} 
	\else
		\usepackage{bibentry}
		\makeatletter\let\saved@bibitem\@bibitem\makeatother
		\usepackage[hidelinks,hypertexnames=false]{hyperref}
		\makeatletter\let\@bibitem\saved@bibitem\makeatother
	\fi
\else
	\ifx\loadbibentry\undefined
		\relax
	\else
		\usepackage{bibentry}
	\fi
\fi

\usepackage[capitalize]{cleveref}
\crefname{equation}{}{}
\Crefname{equation}{}{}
\crefname{claim}{claim}{claims}
\crefname{step}{step}{steps}
\crefname{line}{line}{lines}
\crefname{condition}{condition}{conditions}
\crefname{dmath}{}{}
\crefname{dseries}{}{}
\crefname{dgroup}{}{}

\crefname{Problem}{Problem}{Problems}
\crefformat{Problem}{Problem~(#2#1#3)}
\crefrangeformat{Problem}{Problems~(#3#1#4) to~(#5#2#6)}

\crefname{Theorem}{Theorem}{Theorems}
\crefname{Corollary}{Corollary}{Corollaries}
\crefname{Proposition}{Proposition}{Propositions}
\crefname{Lemma}{Lemma}{Lemmas}
\crefname{Definition}{Definition}{Definitions}
\crefname{Example}{Example}{Examples}
\crefname{Assumption}{Assumption}{Assumptions}
\crefname{Remark}{Remark}{Remarks}
\crefname{Rem}{Remark}{Remarks}
\crefname{remarks}{Remarks}{Remarks}
\crefname{Appendix}{Appendix}{Appendices}
\crefname{Supplement}{Supplement}{Supplements}
\crefname{Exercise}{Exercise}{Exercises}
\crefname{Theorem_A}{Theorem}{Theorems}
\crefname{Corollary_A}{Corollary}{Corollaries}
\crefname{Proposition_A}{Proposition}{Propositions}
\crefname{Lemma_A}{Lemma}{Lemmas}
\crefname{Definition_A}{Definition}{Definitions}

\usepackage{crossreftools}
\ifx\notloadhyperref\undefined
\else
	\relax
\fi

\usepackage{algorithm,algorithmic}

\ifx\loadbreqn\undefined
	\relax
\else
	\usepackage{breqn} 
\fi

\ifx\renewtheorem\undefined
\ifx\useTheoremCounter\undefined
\newtheorem{Theorem}{Theorem}
\newtheorem{Corollary}{Corollary}
\newtheorem{Proposition}{Proposition}

\else

\fi

\fi

\newcommand{\nn}{\nonumber\\ }

\theoremstyle{remark}

\theoremstyle{plain}

\newcommand{\qednew}{\nobreak \ifvmode \relax \else
      \ifdim\lastskip<1.5em \hskip-\lastskip
      \hskip1.5em plus0em minus0.5em \fi \nobreak
      \vrule height0.75em width0.5em depth0.25em\fi}

\NewDocumentCommand{\movedownsub}{e{^_}}{%
  \IfNoValueTF{#1}{%
    \IfNoValueF{#2}{^{}}
  }{%
    ^{#1}
  }%
  \IfNoValueF{#2}{_{#2}}
}

\newcommand{\Real}{\mathbb{R}}

\newcommand{\calC}{\mathcal{C}}

\DeclareSymbolFont{bsfletters}{OT1}{cmss}{bx}{n}
\DeclareSymbolFont{ssfletters}{OT1}{cmss}{m}{n}
\DeclareMathSymbol{\bsfGamma}{0}{bsfletters}{'000}
\DeclareMathSymbol{\ssfGamma}{0}{ssfletters}{'000}
\DeclareMathSymbol{\bsfDelta}{0}{bsfletters}{'001}
\DeclareMathSymbol{\ssfDelta}{0}{ssfletters}{'001}
\DeclareMathSymbol{\bsfTheta}{0}{bsfletters}{'002}
\DeclareMathSymbol{\ssfTheta}{0}{ssfletters}{'002}
\DeclareMathSymbol{\bsfLambda}{0}{bsfletters}{'003}
\DeclareMathSymbol{\ssfLambda}{0}{ssfletters}{'003}
\DeclareMathSymbol{\bsfXi}{0}{bsfletters}{'004}
\DeclareMathSymbol{\ssfXi}{0}{ssfletters}{'004}
\DeclareMathSymbol{\bsfPi}{0}{bsfletters}{'005}
\DeclareMathSymbol{\ssfPi}{0}{ssfletters}{'005}
\DeclareMathSymbol{\bsfSigma}{0}{bsfletters}{'006}
\DeclareMathSymbol{\ssfSigma}{0}{ssfletters}{'006}
\DeclareMathSymbol{\bsfUpsilon}{0}{bsfletters}{'007}
\DeclareMathSymbol{\ssfUpsilon}{0}{ssfletters}{'007}
\DeclareMathSymbol{\bsfPhi}{0}{bsfletters}{'010}
\DeclareMathSymbol{\ssfPhi}{0}{ssfletters}{'010}
\DeclareMathSymbol{\bsfPsi}{0}{bsfletters}{'011}
\DeclareMathSymbol{\ssfPsi}{0}{ssfletters}{'011}
\DeclareMathSymbol{\bsfOmega}{0}{bsfletters}{'012}
\DeclareMathSymbol{\ssfOmega}{0}{ssfletters}{'012}

\DeclarePairedDelimiterX\ip[2]{\langle}{\rangle}{#1,#2}
\DeclarePairedDelimiterX\norm[1]{\lVert}{\rVert}{#1}
\DeclarePairedDelimiterXPP\col[1]{\operatorname{col}}{\{}{\}}{}{#1} 
\DeclarePairedDelimiterXPP\row[1]{\operatorname{row}}{\{}{\}}{}{#1} 
\DeclarePairedDelimiterXPP\erf[1]{\operatorname{erf}}{(}{)}{}{#1}
\DeclarePairedDelimiterXPP\erfc[1]{\operatorname{erfc}}{(}{)}{}{#1}
\DeclarePairedDelimiterXPP\op[2]{\operatorname{#1}}{(}{)}{}{#2} 

\DeclarePairedDelimiterX\Set[2]\{\}{%

#2
}
\DeclarePairedDelimiterX\Setc[1]\{\}{%

#1
}

\NewDocumentCommand\set{s o m}{%
	\IfBooleanTF#1%
	{\IfValueTF{#2}{\Set*{#2}{#3}}{\Setc*{#3}}}%
	{\IfValueTF{#2}{\Set{#2}{#3}}{\Setc{#3}}}%
}

\NewDocumentCommand{\evalat}{s O{\big} m m}{%
  \IfBooleanTF{#1}
   {{\left. #3 \right|_{#4}}}
   {{#3#2|_{#4}}}%
}

\NewDocumentCommand \ifcond {m m} {%
	{#1} %
	\IfValueT{#2}{\, \middle|\, {#2}}%
}

\DeclareDocumentCommand \P {e{_} g >{\SplitArgument{ 1 }{ @| }}d() g } {%
	\mathbb{P}%
	\IfValueTF{#1}{_{#1}}
		{\IfValueT{#2}{_{#2}}}%
	\IfValueT{#3}{\left(\ifcond#3}%
	\IfValueT{#4}{\, \middle|\, {#4}}%
	\IfValueT{#3}{\right)}%
}

\DeclareDocumentCommand \E {e{_} g >{\SplitArgument{ 1 }{ @| }}o g } {%
	\mathbb{E}%
	\IfValueTF{#1}{_{#1}}
		{\IfValueT{#2}{_{#2}}}%
	\IfValueT{#3}{\left[\ifcond#3}%
	\IfValueT{#4}{\, \middle|\, {#4}}%
	\IfValueT{#3}{\right]}%
}

\let\oldforall\forall
\renewcommand{\forall}{\oldforall \, }

\let\oldexist\exists
\renewcommand{\exists}{\oldexist \: }

\graphicspath{{./Figures/}} 
\newcommand{\includeCroppedPdf}[2][]{%
    \IfFileExists{./Figures/#2-crop.pdf}{}{%
        \immediate\write18{pdfcrop ./Figures/#2 ./Figures/#2-crop.pdf}}%
    \includegraphics[#1]{./Figures/#2-crop.pdf}}

\definecolor{gray90}{gray}{0.9}

\ifx\nohighlights\undefined

	\newcommand{\msout}[1]{\text{\color{green} \sout{\ensuremath{#1}}}}
	
	\newcommand{\del}[1]{{\color{green}\ifmmode \msout{#1}\else\sout{#1}\fi}}
\else

	\newcommand{\msout}[1]{#1}
	\newcommand{\del}[1]{#1}
\fi

\newcommand{\hhide}[1]{}

\ifx\diagnoselabel\undefined
	\relax
\else
	\makeatletter
	 \def\@testdef #1#2#3{%
		 \def\reserved@a{#3}\expandafter \ifx \csname #1@#2\endcsname
		\reserved@a  \else
	 \typeout{^^Jlabel #2 changed:^^J%
	 \meaning\reserved@a^^J%
	 \expandafter\meaning\csname #1@#2\endcsname^^J}%
	 \@tempswatrue \fi}
	\makeatother
\fi

\makeatletter
\newcommand*{\addFileDependency}[1]{
  \typeout{(#1)}
  \@addtofilelist{#1}
  \IfFileExists{#1}{}{\typeout{No file #1.}}
}
\makeatother

\usepackage{times}
\title{Building Facade Parsing R-CNN}

\author{Sijie~Wang\thanks{First two authors contributed equally to this work.}$^1$, Qiyu~Kang\footnotemark[1]$^1$, Rui~She$^{1}$, Wee~Peng~Tay$^{1}$,\\  
Diego~Navarro~Navarro$^{2}$, and Andreas~Hartmannsgruber$^{2}$
\thanks{
The authors $^{1}$  are with the Continental-NTU Corporate Lab, 
Nanyang Technological University,
50 Nanyang Avenue, 639798, Singapore.
Emails: { \{ wang1679@e.;qiyu.kang@; rui.she@; wptay@\}ntu.edu.sg}\\%
 The authors $^{2}$  are with Continental Automotive Singapore Pte Ltd,
Emails: { \{diego.navarro.navarro; andreas.hartmannsgruber\}@continental.com}.}%
}

\begin{document}
\maketitle
\begin{abstract}
Building facade parsing, which predicts pixel-level labels for building facades, has applications in computer vision perception for autonomous vehicle (AV) driving. However, instead of a frontal view, an on-board camera of an AV captures a deformed view of the facade of the buildings on both sides of the road the AV is travelling on, due to the camera perspective. We propose Facade R-CNN, which includes a transconv module, generalized bounding box detection, and convex regularization, to perform parsing of deformed facade views. Experiments demonstrate that Facade R-CNN achieves better performance than the current state-of-the-art facade parsing models, which are primarily developed for frontal views. We also publish a new building facade parsing dataset derived from the Oxford RobotCar dataset, which we call the Oxford RobotCar Facade dataset. This dataset contains 500 street-view images from the Oxford RobotCar dataset augmented with accurate annotations of building facade objects. The published dataset is available at \url{https://github.com/sijieaaa/Oxford-RobotCar-Facade} 


\end{abstract}
\section{Introduction}
Building facade parsing or segmentation is a task that classifies the building facade image into elements from different semantic categories. Building facade parsing finds applications in a wide array of fields, including urban augmented reality (AR) \cite{2017FacadeProposal}, camera pose estimation \cite{2021CameraPose}, $3$D building  reconstruction \cite{wu2014building}, and visual Simultaneous Localization And Mapping (SLAM) in street scenes \cite{schops2017large}. The facade segmentation results from general semantic segmentation neural networks \cite{badrinarayanan2017segnet,deeplabv3+,zhao2017pspnet}, although promising, appear coarse. Accurate facade parsing is a challenging task due to the complexity of facade images and the limitation of vanilla semantic segmentation networks that do not incorporate any domain knowledge.

Early learning approaches for building facade parsing like \cite{2011Regionwise} adopt the randomized decision forest and the conditional random field to perform region-wise classification.
The papers \cite{RectilinearParsing2010,2012AutomaticStyleRecognition,2011ShapeGrammar} assume prior knowledge of the regular facade layout or shape grammars for man-made structures to generate neat segmentation maps. 

However, the hand-crafted prior knowledge is highly constrained and these approaches are not robust enough, with a tendency to generate poor predictions in real applications. Recently, with the prevalence of deep learning, convolutional neural networks (CNNs) have been widely adopted to perform building facade parsing.  The work \cite{schmitz2016convolutional}  treats the task as a general image segmentation problem using CNNs without any structural domain knowledge. Later works like DeepFacade \cite{2017DeepFacade} and {PALKN} \cite{2020PyramidALKNet} make use of the regular structure of facades to achieve better semantic results. 
And the work \cite{2018Facadewild} proposes three different network architectures to better dealing with frontal view facade images with varying complexity.

All the above CNN-based facade parsing models like \cite{2017DeepFacade,2020PyramidALKNet} are trained on datasets with approximately frontal facade views. In an autonomous vehicle (AV), an on-board camera typically captures a deformed view of the buildings alongside the road the AV is travelling. See \cref{fig:shear model} for an example. While pre-processing techniques \cite{2021CameraPose} can be used to reconstruct an approximate frontal representation, this additional step can introduce errors and undesirable artifacts, which will subsequently degrade the deep learning model accuracy. 
Alternatively, one can train a current state-of-the-art model like DeepFacade or PALKN using labeled deformed images from an AV camera perspective. However, to the best of our knowledge, there are no datasets containing images from such a camera perspective with accurate building facade annotations.

To better deal with the above challenges, we introduce a dataset with accurate human annotations using the facade images from the public Oxford Radar RobotCar Dataset \cite{RobotCarDatasetIJRR}. This new dataset consists of 500 street-view images, on which we have performed accurate annotations of objects like windows and balconies on building facades. An example is shown in \cref{fig:Dataset Comparison}. 
We further propose a new facade parsing model called Facade R-CNN. Different from general objects, most of the objects on a building facade like windows are highly geometrically constrained. This observation has been similarly utilized in \cite{2017DeepFacade}, where the authors focused on the symmetry properties of facades and proposed regularizers to force the centers of each vertical or horizontal line segment of objects to have small variance. They also proposed to use an object detection module named Faster R-CNN \cite{FasterRCNN} to output bounding boxes (bboxes) for rectangular windows. The main difference to our work is that we directly propose to use the \emph{transconv} module to better learn the symmetric and sheared geometry features of objects. 

We also observe that the objects shown in many facade images like \cref{fig:shear model} are not perfect rectangles. As a result they fit neither the vanilla rectangular bbox nor the symmetric loss constraint in \cite{2017DeepFacade}. We instead propose a less restrictive regularization using the concept of a convex hull, based on the observation that even in a deformed image from an AV camera perspective, objects like windows and doors still maintain convex shapes. Moreover, instead of outputting a rectangular bbox from the object detection module, we propose to output a generalized bbox that regresses a general quadrilateral. 

Our main contributions are summarized as follows:
\begin{enumerate}[1)]
\item We propose Facade R-CNN that consists of a transconv module, generalized bbox detection, and convex regularization to perform facade object recognition from non-frontal building views.
\item We introduce a new facade parsing dataset called the Oxford RobotCar Facade dataset, which contains challenging street-view building facade images captured in an autonomous driving environment and has high-quality annotations.
\item We conduct extensive comparisons and ablation studies to demonstrate that Facade R-CNN achieves the state-of-the-art performance for the facade parsing task.
\end{enumerate}

The rest of this paper is organized as follows.
We present our proposed Facade R-CNN model in  \cref{sec:PROPOSED APPROACH}.  In \cref{sec:Oxford RobotCar Facade Dataset}, we present the Oxford RobotCar Facade dataset. In \cref{sec:Experiments}, we evaluate the performance of our model on three datasets, with comparison to other baseline models. We present further detailed ablation studies in \cref{sec:ablation study}. We conclude the paper in \cref{sec:conclusion}.
We also refer interested readers to the supplementary material for a more detailed account of related works. 

\section{Proposed Approach}\label{sec:PROPOSED APPROACH}
In this section, we present the proposed Facade R-CNN. We introduce three modules: the transconv module, the generalized bbox detection, and the convex regularization. The loss function together with the fusion method is also presented in this section.

\subsection{Network Architecture}
\begin{figure}[!tbp]
    \centering
    \includegraphics[width=0.35\textwidth]{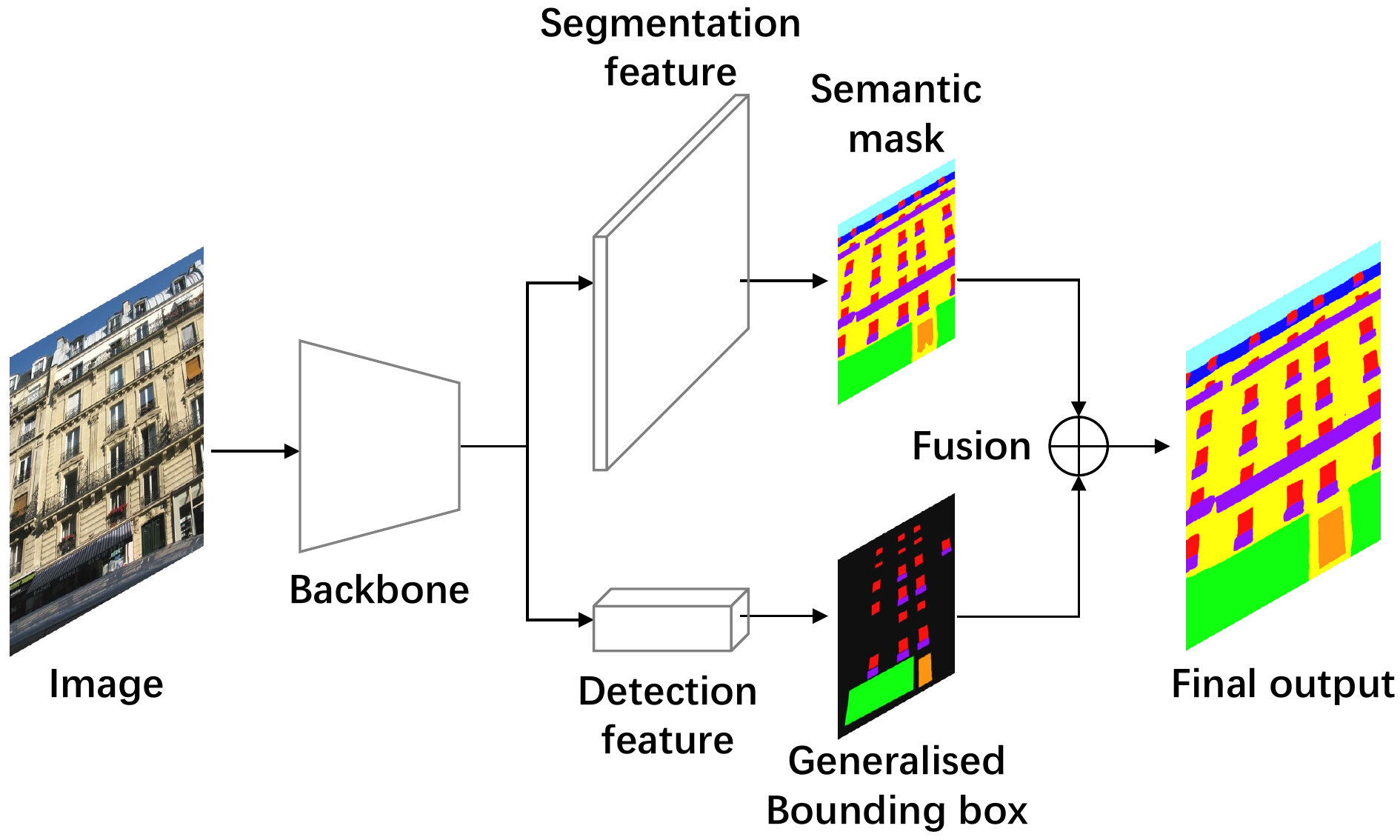}
    \caption{Overview architecture of Facade R-CNN.} 
    \label{fig:Architecture}
\end{figure}

An overview of the architecture of Facade R-CNN is shown in \cref{fig:Architecture}. It is composed of three modules: a backbone, and two branches performing the semantic segmentation decoding and the object detection, respectively. The semantic segmentation decoding branch outputs the pixel-level semantic class for the facade image, while the the object detection branch outputs  object-level \emph{generalized bboxes}, which we further discuss in \cref{subsec:gbbox}. The outputs from the two branches are fused using a score thresholding approach in \cref{subsec:mul learning}.

\subsection{Transconv Module}\label{subsec: transconv module}  

\begin{figure}[!bp]
    \centering
    \includegraphics[scale=0.35]{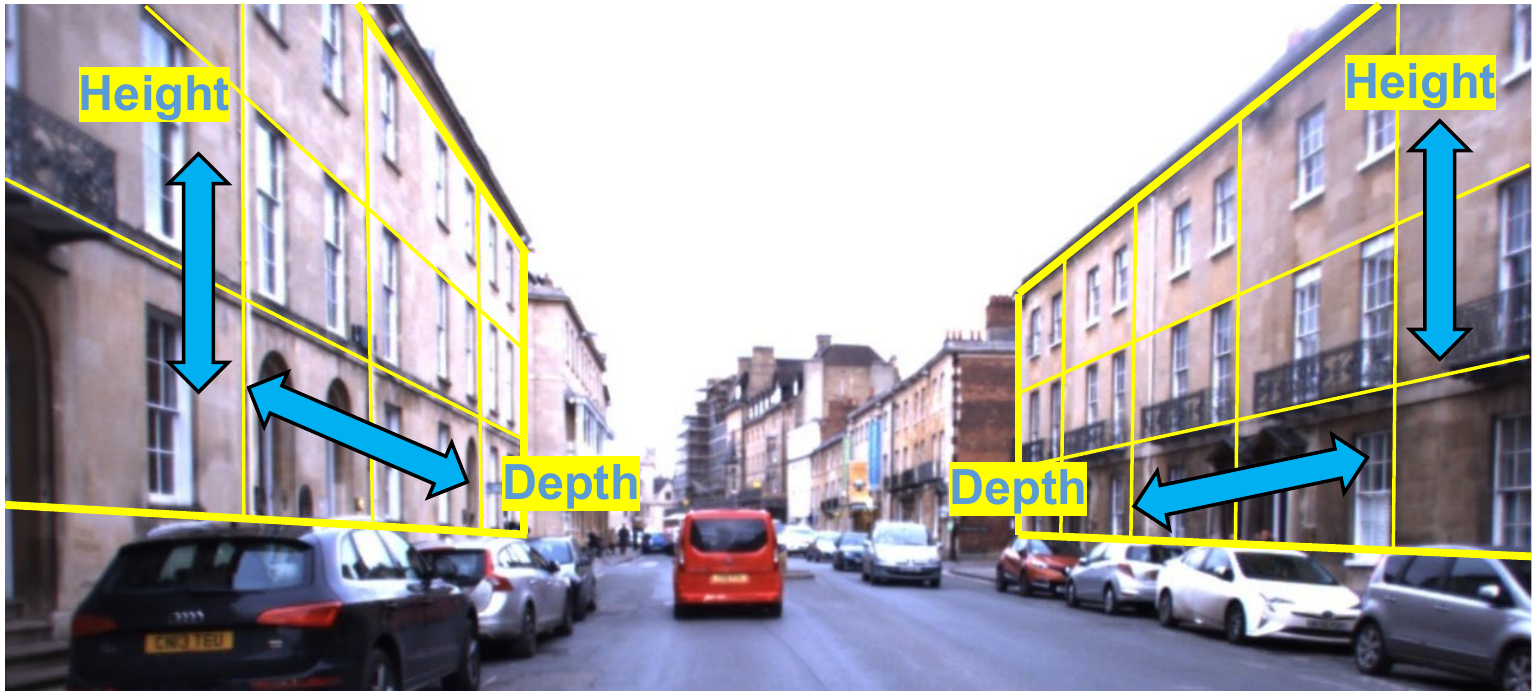}
    \caption{Deformed facade image patterns}
    \label{fig:shear model}
\end{figure}

Unlike objects that have irregular shapes in general segmentation tasks, the objects like windows and balconies on a building facade are highly regular with convex shapes. When the surface of the facade is parallel to the camera imaging plane, such objects in the facade image present as rectangular. However, as alluded to above, in applications like AV driving, buildings may not be oriented parallel to the camera, e.g., as in \cref{fig:shear model}, objects are presented as deformed rectangles in the images. 

We first model the inclined facades as deformed grids shown in \cref{fig:shear model}, with lines corresponding to facade objects' (e.g., windows) edges in both the height and depth directions. We observe that, along the depth direction, the line intersection angles are fixed, while only the scales of the grids on the building facade vary. Using a multi-scaled CNN backbone \cite{ResNet} is robust to scale changes. However, when dealing with the height direction deformations, both the line intersection angles and the scales of the grids are varied, and the multi-scaling strategy is insufficient for convolutional (conv) kernels to detect the deformed patterns. Another latent property of facades along a road is the symmetric distribution. As shown in \cref{fig:shear model}, the left and right building facades are highly symmetric to each other.

According to \cite{krizhevsky2012imagenet}, kernels in the first conv layer play more basic roles to detect geometric patterns like lines and corners. From \cite{GroupCNN}, the combination of rotated conv kernel groups ensures equivariance towards image rotations. We explore solutions to obtain stronger semantics by applying more general affine transformations, including flipping, rotation, and shearing, on the conv kernels to detect more deformed patterns.
Specifically, we propose the \emph{transconv} module to allow the network to parse deformed object images. This module contains a bag of sheared and flipped conv kernels.

Given the vanilla conv kernel, $G_{0}\in{\Real^{W_{G}\times W_{G}\times C_G}}$, we obtain the same sized transformed conv kernel $G_{\mathrm{she},\phi,m}\in{\Real^{W_{G}\times W_{G}\times C_G}}$, where $\phi\in{\left[0^{\circ}, 180^{\circ} \right) }$ are the sheared angles along the second coordinate axis, and $m\in{\{0, 1\}}$ represent the flipping operation, as follows. We first define an intermediate variable:
\begin{align}
G'_{\mathrm{she},\phi,m}(u_{\mathrm{she},\phi,m},v_{\mathrm{she},\phi,m})=G_{0}(u,v),
\end{align}
where $G_{0}(u,v)\in \Real^{C_G}$ is the $(u,v)$-th point in the kernel $G_0$ and $G'_{\mathrm{she},\phi,m}(u_{\mathrm{she},\phi,m},v_{\mathrm{she},\phi,m})$ the corresponding sheared point, is the $(u_{\mathrm{she},\phi,m},v_{\mathrm{she},\phi,m})$-th point in the kernel $G'_{\mathrm{she},\phi,m}$. 
We obtain $(u_{\mathrm{she},\phi,m},v_{\mathrm{she},\phi,m})$ by transforming\footnote{We refer the reader to the supplementary materials for more details about the kernel transformation.} the coordinates $(u,v)$ via: 
\begin{align}
\left[\begin{array}{l}
u_{\mathrm{she},\phi,m} \\
v_{\mathrm{she},\phi,m}
\end{array}\right]
=
\left[\begin{array}{cc}
(-1)^{m} & 0 \\
\tan(\phi) & 1
\end{array}\right] 
\left[\begin{array}{l}
u \\
v
\end{array}\right].
\end{align}

The set of all transformations forms a group with the group  binary operation being the composition of transformations, which we call the \emph{shearing group}. The transformation of conv kernels in our paper is the same as the implementation in \cite[eq. 18]{GroupCNN}.  In \cite{GroupCNN}, the authors proposed to use symmetry groups (subgroups of the isometry group) including $p4$ and $p4m$. By contrast, we propose to use the non-isometry shearing group to better deal with image deformations.

However the above $u_{\mathrm{she},\phi,m}$ and $v_{\mathrm{she},\phi,m}$ are not guaranteed to be integers. We therefore perform an additional bilinear interpolation:
\begin{align*}
G_{\mathrm{she},\phi,m}=\mathrm{itp}(G'_{\mathrm{she},\phi,m}),
\end{align*}
where $\mathrm{itp}(\cdot)$ is the bilinear interpolation function \cite{jaderberg2015spatial}. 

In \cite{GroupCNN}, each group conv layer outputs a set of group feature maps. By contrast, to maintain the succeeding backbone architecture consistency, we perform a summation for the output group features. Given the input $I\in{\Real^{H\times W \times D}}$, the aggregated feature  $I^{\prime}\in{\Real^{H^{\prime}\times W^{\prime} \times D^{\prime}}}$ is obtained via: 
\begin{align}
I^{\prime}= \sum_{\phi,m}{G_{\mathrm{she},\phi,m} * I},\label{eq:conv}
\end{align}
where $*$ denotes the convolution operation. 
By incorporating features under transformed conv kernels, the transconv module can detect more deformed patterns and thus is more robust when dealing with facade images from on-board camera views.



\subsection{Generalized Bounding Box Detection}\label{subsec:gbbox}
\begin{figure}[htbp]\footnotesize
    \centering
    \includegraphics[width=0.25\textwidth]{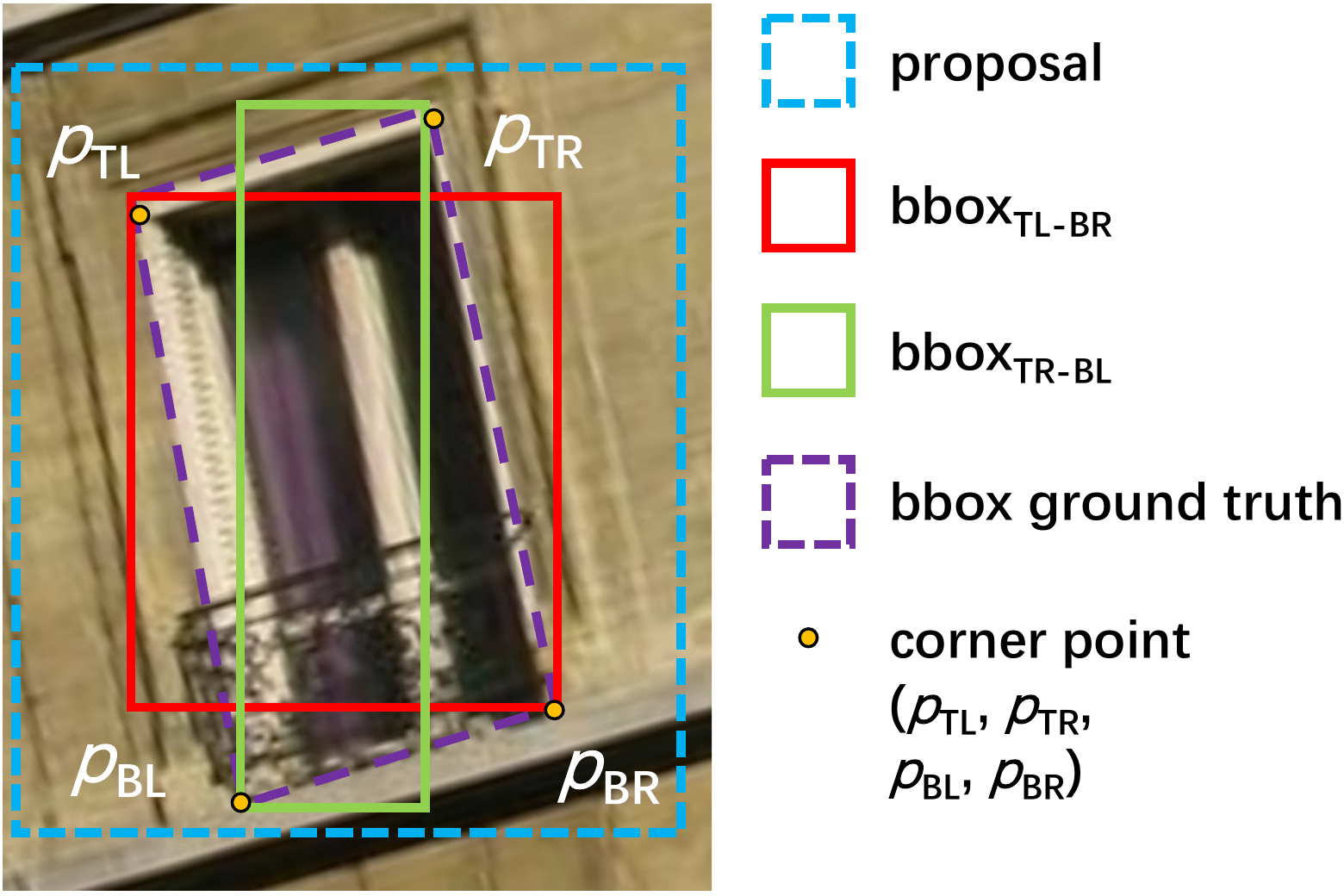}
    \caption{Generalized bounding box.}
    \label{fig:generalized bbox illustration}
\end{figure}

\begin{table}[!htb]\footnotesize
\centering
\begin{tabular}{l c c c} 
\toprule
Head & MAdd & Flops & Memory \\
\midrule
Mask R-CNN & 1.06G & 478.17M & 1.56M \\
Faster R-CNN & 27.87M & 13.93M & 0.01M \\
Facade R-CNN & 27.93M & 13.97M  & 0.01M \\
\bottomrule
\end{tabular}
\caption{MAdd, Flops, and memory usage statistics comparison for different R-CNN heads.}
\label{tab:network params statistics}
\end{table}

In a typical object detection network \cite{FasterRCNN}, the output is a rectangular bbox. In DeepFacade \cite{2017DeepFacade}, the vanilla rectangular bbox is used to refine the rectangular objects like windows on the building facade. However, the rectangular bbox is only applicable to rectified facade images taken from the frontal perceptive which is rare in actual situations like images taken from on-vehicle front cameras. As illustrated in \cref{subsec: transconv module}, when the camera view is changed, the objects on the facade no longer have a rectangular shape, while a rectangular bbox would still regard the deformed quadrilateral as the rectangle and hence result in wrong predictions for some pixels.

To construct a more robust object detection head to refine the output from the semantic segmentation branch in \cref{fig:Architecture}, we propose the \emph{generalized bbox detector}, which can better fit the deformed facade objects. As shown in \cref{fig:generalized bbox illustration}, we first define the top-left, top-right, bottom-left and bottom-right corners of the window to be $p_\mathrm{TL}$, $p_\mathrm{TR}$, $p_\mathrm{BL}$, and $p_\mathrm{BR}$, respectively. Then, for a general quadrilateral object, we use two bboxes to represent it: the bbox$_{\mathrm{TL-BR}}$ formed by $p_\mathrm{TL}$ and $p_\mathrm{BR}$, and the bbox$_\mathrm{TR-BL}$ formed by $p_\mathrm{TR}$ and $p_\mathrm{BL}$. The two rectangular bboxes are used respectively to find the two sets of non-adjacent vertices of the quadrilateral object.

\begin{figure}[!htbp]
    \centering
    \includegraphics[scale=0.27]{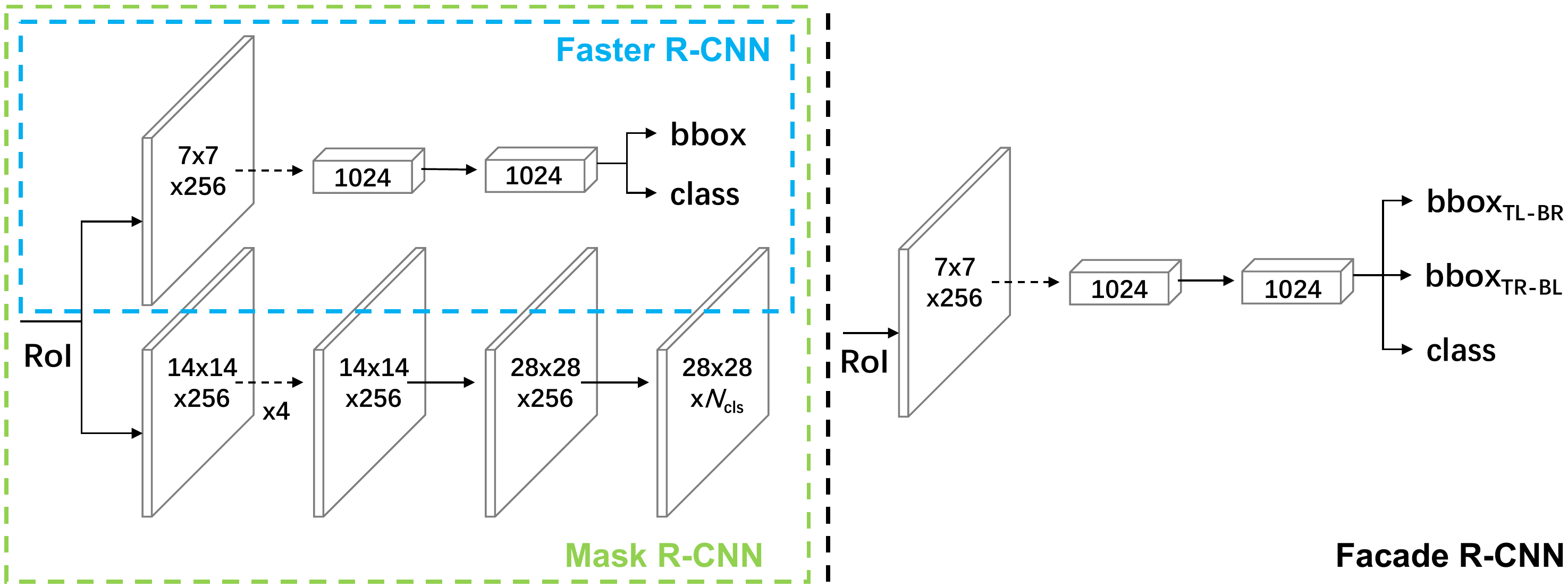}
    \caption{Head comparison. Left: the heads of Faster R-CNN and Mask R-CNN. Right: the head of our Facade R-CNN.}
    \label{fig:generalized bbox detection head}
\end{figure}

We construct the generalized bbox detection head as shown in \cref{fig:generalized bbox detection head}, which is based on the basic Faster R-CNN head. The Mask R-CNN predicts the dense semantic map by adding an extra FCN branch. By contrast, our Facade R-CNN that specializes to facade object parsing does not require any extra segmentation module. To demonstrate the design efficiency for our generalized bbox detection head, we show head computation statistics in \cref{tab:network params statistics}\footnote{We use torchstat from \url{https://github.com/Swall0w/torchstat} as the analysis tool.}.
Compared with the Mask R-CNN head, our pure bbox regression head consumes 1/30 less MAdd and Flops and 1/150 less memory usage, and has similar efficiency as Faster R-CNN.

\begin{figure}[!htbp]
    \centering
    \includegraphics[width=0.35\textwidth]{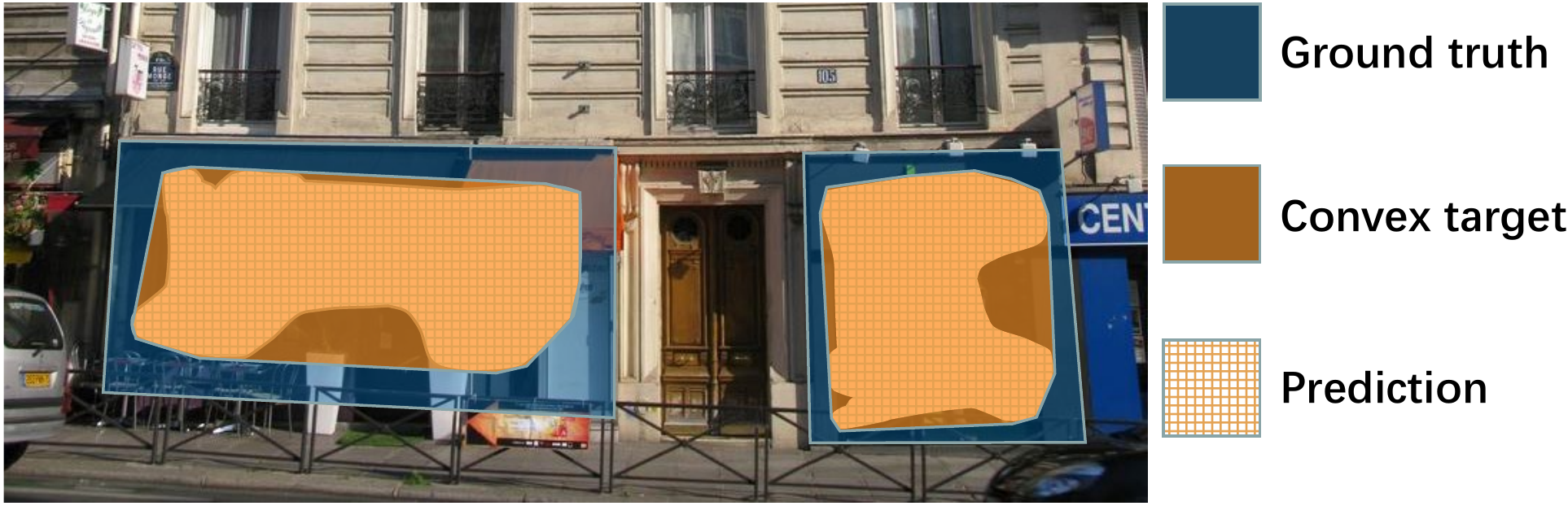}
    \caption{Convex regularization for the non-convex segmentation output.}
    \label{fig:convex shop}
\end{figure}

Following Faster R-CNN, we first use the Region Proposal Network (RPN) to generate candidate proposals. In the Faster R-CNN, each proposal has only one rectangular bbox regression target. By contrast, in our Facade R-CNN, each proposal regresses two rectangular bboxes to construct the final generalized bbox as illustrated above. We minimize the following objective function:
\begin{align}
    L_{\mathrm{detection}} = \frac{2}{N_{\mathrm{bbox}}}\bigg( \sum_{k=1}^{N_{\mathrm{bbox}}/2} \big(\ell_{\mathrm{ce}} (t_{\mathrm{cls},k}, t_{\mathrm{cls},k}^{*}) \nn
    + \sum_{i=1}^2 \ell_{\mathrm{smoothL1}}(t_{xywh,k,i} - t_{xywh,k,i}^{*}) \big) \bigg),
\end{align}
where $N_{\mathrm{bbox}}$ is the number of output rectangular bboxes, 
$t_{\mathrm{cls},k}$ is a probability vector for classification of the $k$-th generalized bbox,
$t_{\mathrm{cls},k}^{*}$ is the classification target, $t_{xywh,k,i}$ is a vector that contains center coordinates, width, and height outputs of the $i$-th ($i\in\{1,2\}$) rectangular bbox for the $k$-th generalized bbox, $t_{xywh,k,i}^{*}$ is the regression target, $\ell_{\mathrm{ce}}$ is the cross-entropy loss, and $\ell_{\mathrm{smoothL1}}$ is the smooth $L_1$ function \cite{girshick2015fast}.

\subsection{Convex Regularization}
In this section, we introduce a convex regularization due to the observation that in the context of building facade parsing, many objects like windows, shops and doors are all shown as deformed rectangles in images taken from different view perspectives. The ground truth instance (e.g., an individual window or door) segmentation masks over those objects therefore present general convex shapes. However as we observe in \cref{fig:convex shop}, the segmentation predictions from semantic segmentation networks like FCN \cite{2015FCN} are in non-convex irregular shapes for the two shops. We therefore propose a convex regularization to improve the robustness of the network and allow the network to maximally extract information from a convex region. In each training iteration, we additionally \emph{dynamically} generate convex masks, called \emph{convex target}, as the extra targets to guide the model training.

Specifically, given the set of pixels, denoted as $S_{i}$, in which each pixel is predicted as the $i$-th class, and the ground truth segmentation labels, our convex target mask $S_{\mathrm{cvx},i}^{*}$ is obtained by:
\begin{align}
    S_{\mathrm{cvx},i}^{*} = \bigcup_{k=1}^{N_{\mathrm{cls},i}^{*}} \bigg( \Gamma \bigg( S_{i} \bigcap S_{i,k}^{*} \bigg) \bigg),\label{eq:cvx}
\end{align}
where $N_{\mathrm{cls},i}^{*}$ is the number of  ground truth instance  mask of the $i$-th class; $S_{i,k}^{*}$ is the $k$-th ground truth instance mask of the $i$-th class and $\Gamma(\cdot)$ is the convex hull of its set argument. The instance masks can be generated using connected components labeling.
We then compute the convex regularizer as:
\begin{align}
    L_{\mathrm{cvx}} = \frac{1}{|\calC_{\mathrm{cls}}|} \sum_{i \in \calC_{\mathrm{cls}}} \ell_{ce} (S_{\mathrm{cvx},i}^{*}),
\end{align}
where $\calC_{\mathrm{cls}}$ is set of classes that have convex mask shapes, e.g., windows, shops and doors, and $\ell_{ce}(S)$ is the pixel-wise cross-entropy loss between predictions and labels restricted to the pixel set $S$.

\subsection{Multi-task Learning}\label{subsec:mul learning}
Our proposed Facade R-CNN consists of two branches:\footnote{See supplementary materials for more details.} segmentation branch and detection branch. We adopt the DeepLabV3 \cite{deeplabv3} as the base network for semantic parsing, while the detection head illustrated in \cref{subsec:gbbox} is used for generalized bbox refinement.

In the training stage, the overall loss function is defined as:
\begin{align}
L=L_{\mathrm{semantic}} + L_{\mathrm{proposal}} + L_{\mathrm{detection}} + \alpha L_{\mathrm{cvx}},\label{eq:finalloss}
\end{align}
where $L_{\mathrm{semantic}}$ is the cross-entropy semantic segmentation loss, $L_{\mathrm{proposal}}$ is the RPN loss function defined in \cite{FasterRCNN}, and $\alpha$ is a pre-defined weight of the convex regularization.

A mask fusion strategy is critical for generating the refined facade parsing output. One way is to directly perform union over the mask outputs from the semantic and detection branches. This however inevitably introduces unpredictable errors, as some generalized bboxes are inaccurate. Thus we propose to apply a score threshold for the detected bboxes. As illustrated in \cref{subsec:gbbox}, each output generalized bbox is associated with a classification score $s_{k}=\max(t_{\mathrm{cls},k})\in \left[ 0,1 \right]$. In the testing stage, after obtaining the generalized bbox and the semantic segmentation prediction, for the pixels in the overlapping region of these two outputs, the final fused semantic output for pixel $j$ is generated as follows:
\begin{align}
Y_{j}=\begin{cases}
D_{j} ,&  s_{j}>T, \\
S_{j} ,&  \text{otherwise,}
\end{cases}\label{eq:infer}
\end{align}
where  $T$ is a pre-defined generalized bbox score threshold, $D_j$ is the segmentation class predicted by the generalized bboxes at pixel $j$, and $S_j$ is the segmentation class generated from the semantic branch at pixel $j$.

\section{Oxford RobotCar Facade Dataset}\label{sec:Oxford RobotCar Facade Dataset}
In this section, we briefly describe the characteristics of the dataset, Oxford RobotCar Facade. We refer the readers to the supplementary material for more details of the new dataset.

\begin{figure}[!htbp]
    \centering
    \includegraphics[width=0.47\textwidth]{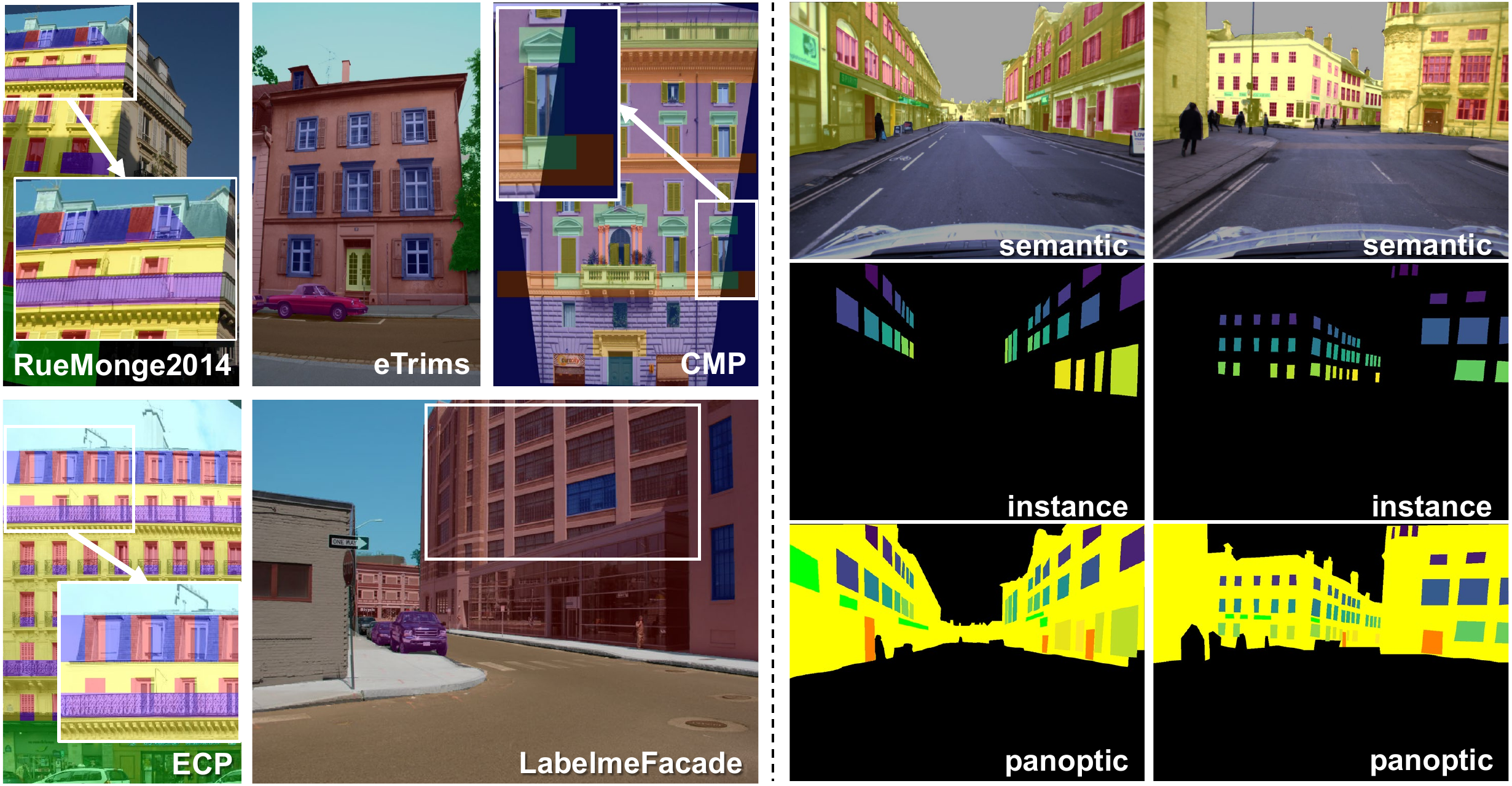}
    \caption{Dataset comparison. Left: existing building facade parsing datasets, where the inaccurate annotations are highlighted with boxes; right: our proposed dataset.}
    \label{fig:Dataset Comparison}
\end{figure}

To the best of our knowledge, the Oxford RobotCar Facade dataset is the first driving environment facade dataset made publicly available. There are $500$ images, each of size $1280\times 960$ and high-quality 5-class annotations: \emph{window}, \emph{door}, \emph{balcony}, \emph{shop}, and \emph{facade}. Some samples are shown in \cref{fig:Dataset Comparison}. The whole dataset is based on the large autonomous driving dataset Oxford Radar RobotCar dataset \cite{RobotCarDatasetIJRR}, which is collected along a consistent route through Oxford, UK. As a comparison, many other existing facade datasets such as the RueMonge2014 \cite{Monge2014} and ECP \cite{ECP} either have inaccurate annotations or less than $250$ annotated images.  We list the features of Oxford RobotCar Facade dataset as follows. 

\textbf{Accurate annotations.}
To ensure the quality of annotations, we manually label all classes based on the Labelme tool\footnote{See \url{https://github.com/wkentaro/labelme}}, instead of using inaccurate pre-defined grammars to intuitively accelerate the labeling process. Also, we carefully deal with the problem of occlusion, i.e., we specifically give different labels to distinguish the foreground objects (e.g., vehicle and pedestrians) from background facades. 

\textbf{Diversity.}
The 500 images we selected consists of various buildings, such as churches, apartments, and office buildings, which largely increase the sample diversity. In addition, since the scenes are captured on a consistent route, the same building facade would have different shapes from varying camera views. Therefore, our dataset is more challenging, which can thus better reflect the generalization ability of parsing models.

\textbf{Multi-task support.}
To build a comprehensive benchmark for building facade parsing, we provide not only semantic annotations, but also instance and panoptic annotations \cite{kirillov2019panoptic}. Specifically, we use the object-level class \emph{window} for instance segmentation task, while the other 4 stuff-level classes \emph{facade}, \emph{door}, \emph{balcony}, \emph{shop} together with \textit{window} are used in the panoptic segmentation task. For a fair benchmark comparison, we split the whole dataset into training (400 images) and testing (100 images) subsets.

\section{Experiments}\label{sec:Experiments}

We evaluate our proposed approach and several baselines on three datasets, the ECP dataset \cite{ECP}, the RueMonge2014 dataset \cite{Monge2014}, and our Oxford RobotCar Facade dataset. We select four general segmentation CNNs and two state-of-the-art building facade parsing networks as baseline models, including FCN \cite{2015FCN}, PSPNet \cite{zhao2017pspnet}, DeepLabV3 \cite{deeplabv3}, DeepLabV3+ \cite{deeplabv3+}, DeepFacade \cite{2017DeepFacade}, and Pyramid ALKNet \cite{2020PyramidALKNet}

\subsection{Dataset and Training Configuration}\label{subsec:dataset configuration}
In all the experiments except those in \cref{sec:ablation study}, we use the loss function defined in \cref{eq:finalloss} with $\alpha=1/9$ and the fusion strategy in \cref{eq:infer} with $T=0.5$.  
We refer the reader to the the supplementary material for more details of the datasets with specific settings and the training configurations. 


\subsection{Main Results}\label{subsec:Results}
The comparisons between our model and baselines on all datasets are shown in \cref{tab:Results on the datasets}. On the ECP dataset, though we obtain slightly lower mIoU compared with PALKN, we still surpass all the counterparts in accuracy. On the RueMonge2014 dataset, we clearly observe that Facade R-CNN outperforms all the other models and obtain the highest scores, $74.34$ in mIoU and $88.67$ in accuracy. Compared with DeepLabV3, our proposed model shows $+1.06$ and $+0.38$ respective improvements in the two metrics. Also, our model surpasses the previous state-of-the-art facade parsing network PALKN by $+0.92$ and $+0.27$.


\begin{table}[!tb]\footnotesize
\centering
\begin{tabular}{l c c c c c c} 
\toprule
\multirow{2}{*}{Model} & \multicolumn{2}{c}{ECP} & \multicolumn{2}{c}{RueMonge} & \multicolumn{2}{c}{Oxford} \\
 & mIoU & Acc. & mIoU & Acc. & mIoU & Acc.\\
\midrule
FCN & 84.4 & 93.31 &    72.85 & 88.18 & 45.12 & 93.99 \\
PSPNet & 83.78 & 93.17 &    70.93 & 87.24 & 48.87 & 94.05 \\
V3 & 83.76 & 93.4 &            73.28 & 88.29 & 51.3 & 94.38 \\
V3+ & 84.29 & 93.35 &  72.96 & 88.09 & 50.33 & 94.57 \\
\midrule
DFacade & 83.78 & 93.54 & 71.12 & 87.33 & 47.31 & 94.49  \\
PALKN & \textbf{84.9} & 93.56 & 73.42 & 88.4 & 51.22 & 94.61 \\
Ours & 84.47 & \textbf{93.78} &  \textbf{74.34} & \textbf{88.67} & \textbf{53.8} & \textbf{94.67} \\
\bottomrule
\end{tabular}
\caption{Results on the three benchmark datasets.}
\label{tab:Results on the datasets}
\end{table}

Our proposed Oxford RobotCar Facade dataset is a challenging dataset, where the images are captured in noisy driving environments. As shown in \cref{tab:Results on the datasets}, Facade R-CNN outperforms all the baselines, and achieves the highest mIoU of $53.8$ and the highest accuracy of $94.67$. Compared with PALKN, Facade R-CNN shows improvements of $+2.58$ in mIoU. Meanwhile, our three proposed modules bring $+2.5$ gain over the base network DeepLabV3. The comparison demonstrates that our model is better at dealing with challenging situations and has stronger robustness to the distortion caused by camera view change.


\section{Ablation Study}\label{sec:ablation study}
To better evaluate the proposed transconv module, generalized bounding box detection, and convex regularization, we conduct extensive ablation experiments. We \emph{individually} add each module to the baseline DeepLabV3 to conduct experiments. For a fair comparison, we do not leverage data augmentation tricks in this section as they would introduce uncertainties in the conclusions.  

\subsection{Transconv Module}\label{subsec: Ablation study on the transconv module}
In this part, we first analyze the performance of different combinations of affine transformation in the first conv layer. From \cref{tab:Ablation study on the transconv module}, we observe that combining shearing and flipping together is the optimal strategy, which contributes $+0.65$ mIoU improvement to the baseline. Applying the transconv module on the first residual stage can also bring $+0.25$ gain in accuracy. However, when the transconv module is added into the succeeding deeper layers, it is not useful anymore, which is consistent with the illustration in \cref{subsec: transconv module} that the first few conv layers are more able in detecting basic geometry patterns.

\begin{table}[!tb]\footnotesize
    \centering
    \begin{tabular}{l  c c c c   c} 
\toprule
Depth & Shear & Flip & Rotate  & mIoU & Accuracy \\
\midrule
\multirow{6}{*}{first layer}
 & & & &  70.37 & 87.5\\
 & \checkmark & & & 70.61 & 87.51 \\
 & & \checkmark & &  70.45 & 87.64 \\
 & & &\checkmark &  70.47 & 87.48 \\
 &\checkmark & \checkmark & &  \textbf{71.02} & 87.57 \\
 &\checkmark & \checkmark & \checkmark &  70.79 & 87.52 \\
\midrule
+ stage1 &\checkmark & \checkmark &  & 70.41 & \textbf{87.75} \\
+ stage2 &\checkmark & \checkmark &  & 69.66 & 87.41 \\
+ stage3 &\checkmark & \checkmark &  & 69.73 & 87.28 \\
+ stage4 &\checkmark & \checkmark &  & 67.58 & 86.53 \\
\bottomrule
\end{tabular}
\caption{Comparisons of transformations on the RueMonge2014 dataset. For the first conv layer, we replace the first 7$\times$7 conv kernel; for later stages, we replace the middle 3$\times$3 conv kernel in each residual block.}
\label{tab:Ablation study on the transconv module}
\end{table}

\subsection{Generalized Bounding Box Detection}\label{subsec:exp_gbbox}
In \cref{tab:Ablation study ge bbox on monge}, we first test different threshold $T$ setting from $0$ to $0.9$ for mask fusion, where neither too high nor too low value can obtain significant gain. By contrast, the optimal threshold $0.5$ shows the optimal performance that increases mIoU by $+0.39$. 


\subsection{Convex regularization}\label{subsec:convex regularization}
\begin{center} 
\begin{figure}[hbtp]
    \centering
    \includegraphics[width=0.4\textwidth]{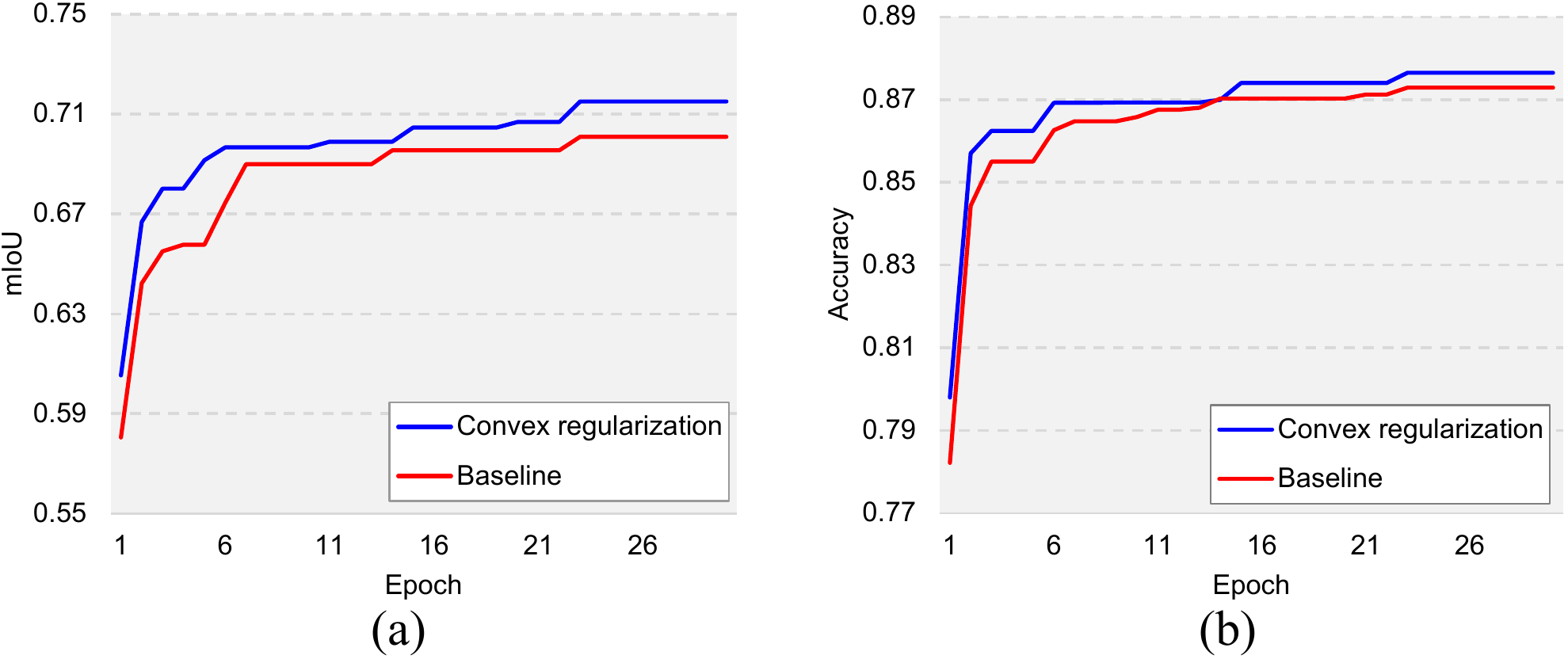}
    \vspace{-0.2cm}
    \caption{The network optimal performance with convex regularization. (a) mIoU \emph{vs.} epoch; (b) accuracy \emph{vs.} epoch. }
    \label{fig:convergence graph}
\end{figure}
\end{center}

The convex regularizer serves as an extra loss for efficient network convergence. We explore different weight settings for the convex regularizer as shown in \cref{tab:ablation study for convex weight}. From \cref{tab:ablation study for convex weight}, we observe the optimal weight value of $1/9$ achieves the best result, which outperforms the baseline by $+0.83$ mIoU and $+0.15$ accuracy. We also evaluate the performance when adding extra weight for convex classes, i.e., if we fix $ S_{\mathrm{cvx},i}^{*}=S_{i,k}^{*}$ instead of using \cref{eq:cvx}, we do not get significant improvement.
We visualize the network convergence with convex regularization in \cref{fig:convergence graph}, which shows both accuracy and mIoU can converge faster than the baseline counterpart.

\begin{table}[!htb]\footnotesize
\centering
\begin{tabular}{lcccc}
\toprule
Weight & \multicolumn{2}{c}{Convex regularization} & \multicolumn{2}{c}{Extra loss weight} \\
& mIoU & Accuracy & mIoU & Accuracy\\
\midrule
Baseline & 70.37 & 87.5 & - & - \\
1/12 & 70.54 & 87.44  & 70.6 & 87.51\\
1/9 & \textbf{71.2} & \textbf{87.65} & 70.12 & 87.38 \\
1/6 & 70.66 & 87.57 & 70.55 & 87.56\\
1/3 & 70 & 87.19 & 70.6 & 87.63\\
\bottomrule
\end{tabular}
\caption{Convex regularization performance under different weight settings on the RueMonge2014 dataset.}
\label{tab:ablation study for convex weight}
\end{table}

 \vspace{-0.7cm}
\subsection{Generalization}
After ablating the three proposed modules, we further inspect the generalization performance of our network. In this section, we use the combination of both the ECP and RueMonge2014 as the training dataset, while the Oxford RobotCar Facade is set as the testing dataset. As shown in \cref{tab:Results using ecpmonge}, our Facade R-CNN outperforms all counterparts even further by at least $+1.14$ in mIoU and $+2.06$ in accuracy, which demonstrates that our model is more able to generalize parsing ability to more challenging datasets.


\begin{table}[!htb]\footnotesize
\centering
\begin{tabular}{lcc}
\toprule 
Threshold & mIoU & Acc. \\
\midrule
Baseline & 70.37 & 87.50 \\
0.9 & 70.46 & 87.53 \\
0.7 & 70.48 & 87.50 \\
0.5 & \textbf{70.76} & \textbf{87.60} \\
0.3 & 69.87 & 87.42 \\
0.1 & 69.80 & 87.30 \\
0 & 69.32 & 86.99 \\
\bottomrule
\end{tabular}
\caption{Comparison of the threshold settings for generalized bbox fusion on the RueMonge2014 dataset.}
\label{tab:Ablation study ge bbox on monge}
\end{table}

\begin{table}[!htb]\footnotesize
\centering
\begin{tabular}{lcc} 
\toprule
Model & mIoU & Acc. \\
\midrule
FCN & 22.91 & 71.68\\
V3 & 24.06 & 74.10 \\
V3+ & 23.20 & 75.24\\
\midrule
PALKN & 23.77 & 74.80 \\
Ours  &  \textbf{25.20} &  \textbf{77.30}\\
\bottomrule 
\end{tabular}
\caption{Generalization performance.}
\label{tab:Results using ecpmonge}
\end{table}

\section{Conclusion}\label{sec:conclusion}
In this paper, we investigate the problem of building facade parsing in realistic street-view scenes where building facade images are from non-frontal perspectives. To achieve the goal, we propose the transconv module, generalized bbox detection, and convex regularization, all of which form the Facade R-CNN. Extensive experiments are conducted to compare the our model with other baselines. We demonstrate that the proposed Facade R-CNN achieve the state-of-the-art performance. To advance the application of building facade parsing in autonomous driving, we publish a new dataset Oxford RobotCar Facade, which has realistic street-view images and high-quality annotations.


\appendix
\section{Related Work}\label{sec:relatedworks}
{In this section we brief more works that deal with the building facade parsing task from both the traditional and deep learning communities.

\textbf{Traditional Building Facade Parsing.} Traditional approaches tackle the problem of building facade parsing by focusing on hand-crafted prior knowledge. In \cite{RectilinearParsing2010}, buildings are parsed as individual facades. Each facade is formatted as the combination of roof and floor, and a dynamic optimization is then applied. The work \cite{2012AutomaticStyleRecognition} first splits the whole street-view buildings into facades.  Hand-crafted features are then extracted based on these facades, which are finally used for building style recognition.
The reference \cite{2011ShapeGrammar} formulates the problem as a hierarchical Markov decision process, where a binary split grammar is applied to parse and obtain the model optimal facade layout.

\textbf{Deep Learning Segmentation. }
CNNs have shown the powerful ability on vision-based tasks, including the classification, the object detection, and the semantic segmentation. Various CNN-based models are proposed to tackle computer vision problems. Fully Convolutional Network (FCN)\cite{2015FCN} is a milestone in image segmentation, which regards the semantic segmentation as a pixel-wise classification task.
In \cite{unet},  U-Net is proposed mainly for medical image segmentation to capture more detailed representation. Unlike FCN that excludes information of the first several layers, U-Net integrates features from all convolutional (conv) layers by skip connections. The PSPNet \cite{zhao2017pspnet} introduces the pyramid pooling module which extracts features from multi-scales with multiple pooling operations, and the final prediction is generated based on the concatenated features. This module aggregates context information from a wider field, which demonstrates to be helpful to deal with scene parsing task. The {DeepLab} series \cite{deeplabv3,deeplabv3+}  enlarge the receptive field of conv layers by leveraging the dilated conv \cite{yu2016dilatedconv}. The dilated conv does not limit the kernel to have successive pixels anymore. By introducing the dilated rate, dilated conv enables pixels in the kernel to be far away from each another, without largely increasing the model parameters.}

\textbf{Deep Learning Building Facade Parsing.} The work \cite{schmitz2016convolutional} is the pioneer that introduces CNN into the facade parsing task. It proposed a network that is based on AlexNet \cite{deng2009imagenet}, where the encoding part consists of five conv layers while the decoding part consists of two conv layers and four fully-connected layers.
Deepfacade \cite{2017DeepFacade} first combines semantic segmentation and object detection together to obtain better building facade parsing result. It uses the rectangular bounding box (bbox) produced by Faster R-CNN to better regress the segmentation output. In addition, it utilizes a symmetry loss function that leverages the symmetric information of facade objects like windows. 
The work {PALKN} {\cite{2020PyramidALKNet}  follows the success of dilated conv. It tackles the problem of occlusion and ambiguous of facades by introducing the atrous large kernel module (ALK module). With the proposed ALK module, their network is able to recognize patterns in a wider field and make use of the regular structures of facades to aggregate useful non-local context information.}

\section{Transconv Module Details}
\begin{center} 
\begin{figure}[htbp]
    \centering
    \includegraphics[width=0.45\textwidth]{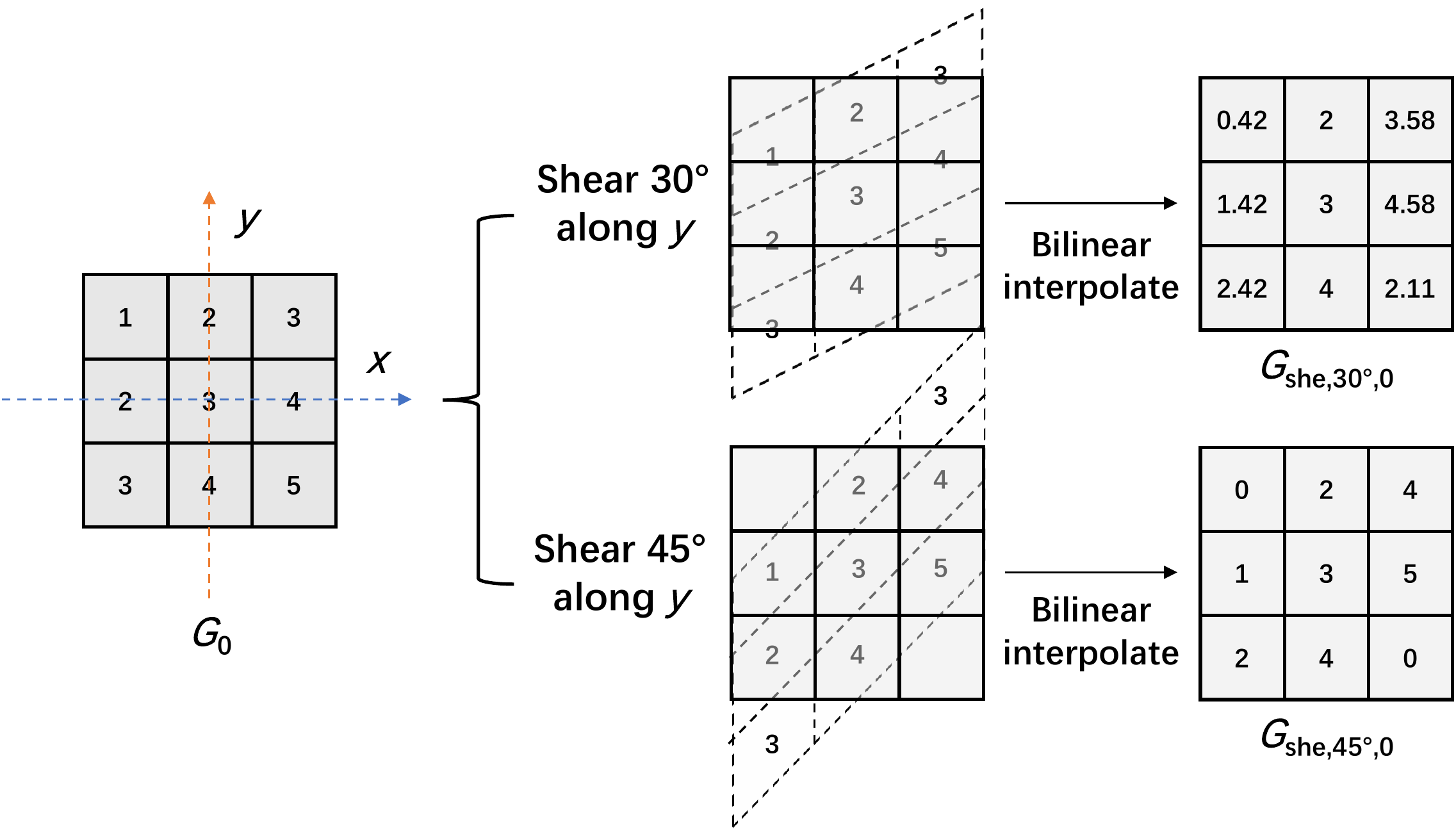}
    \caption{Illustration for transforming the vanilla conv kernel to obtain the sheared conv kernel.} 
    \label{fig supp:transconv}
\end{figure}
\end{center}

Given the vanilla conv kernel $G_{0}$, we first apply shearing to obtain the sheared grid. Then, the bilinear interpolation is applied to generate the neat grid which is the final sheared kernel $G_{\mathrm{she},\phi,m}$. An example is shown in \cref{fig supp:transconv}, where two sheared kernels $G_{\mathrm{she},30^{\circ},0}$ and $G_{\mathrm{she},45^{\circ},0}$ with shearing angles 30$^{\circ}$ and 45$^{\circ}$ along the \emph{y}-axis are obtained from the vanilla kernel $G_{0}$.

\section{Oxford RobotCar Facade Dataset}\label{sec:Oxford RobotCar Facade Dataset}
\begin{center} 
\begin{figure}[htbp]
    \centering
    \includegraphics[width=0.45\textwidth]{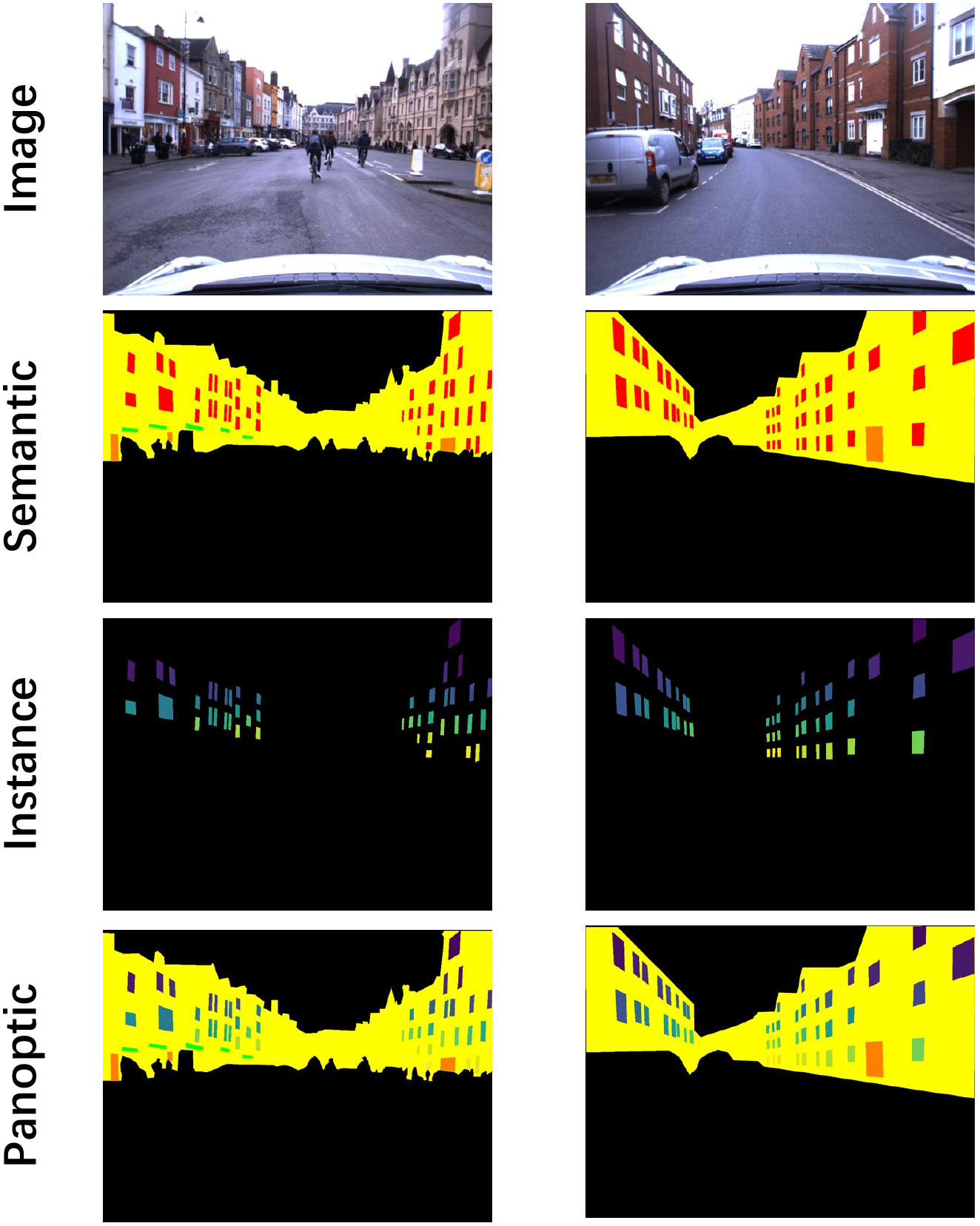}
    \caption{The Oxford RobotCar Facade dataset.} 
    \label{fig supp:oxford}
\end{figure}
\end{center}

To support the building facade parsing community, many facade datasets have been proposed during the last several years. The RueMonge2014 \cite{Monge2014} is a facade dataset collected along the Rue Monge street in Paris, which contains both 3D and 2D semantic annotations. As the 2D semantic labels are generated based on the 3D building models, there exists some mismatching between labels and images as shown in \cref{fig:Dataset Comparison} in the paper. The eTrims \cite{etrims} incorporates buildings in various environments with different views and provides highly-accurate annotations. However, this dataset only contains a total of 60 images, which is inadequate for model evaluation. The CMP and the ECP datasets \cite{CMP,ECP} contain rectified building facades of 378 and 104 images respectively. However, these two datasets both intuitively regard the facade as axis-aligned rectangular layout, which inevitably leads to label-image mismatching, as shown in \cref{fig:Dataset Comparison} in the paper. LabelmeFacade \cite{LablemeFacade} is a large facade dataset that collects 945 facade images in different views. It has rich annotations for road-side objects, such as trees and vehicles. Nevertheless, it does not provide facade object annotations in a unified level, i.e., they only annotate part of the windows and doors on the facade while leaving the rest unlabeled as shown in \cref{fig:Dataset Comparison} in the paper. This would cause misleading during training stage and finally affect the performance of parsing models. The more detailed visualization of our dataset is shown in \cref{fig supp:oxford}.

\begin{table}[]\footnotesize
\centering
\begin{tabular}{c c c c} 
\toprule
Class  & Images no. & Pixels no. & Instances no. \\
\midrule
\textit{facade} & 500 & 14109k & - \\
\textit{window} & 500 & 1868k & 8820 \\
\textit{door} & 214 & 151k & - \\
\textit{balcony} & 52 & 38k & - \\
\textit{shop} & 74 & 34k & - \\
\bottomrule 
\end{tabular}
\caption{The statistics of the data samples in our dataset.}
\label{tab:Oxford RobotCar Facade}
\end{table}

\section{Dataset and Training Configuration}\label{subsec:dataset configuration}
\subsection{Dataset Configuration}
We introduce the datasets used in our experiments with the specific settings as follows. 

\textbf{ECP.}\
The ECP dataset contains a total of 104 rectified images. We use the improved annotations provided by Mathias et al. \cite{ECPAnnotations}, where there are 8 classes:
\emph{window}, \emph{door}, \emph{balcony}, \emph{shop}, \emph{facade}, \emph{sky}, \emph{roof}, \emph{chimney}.
We preform five-fold cross validation on this dataset same as the paper \cite{2020PyramidALKNet} .

\textbf{RueMonge2014.}\
The RueMonge2014 dataset contains a total of 219 deformed images with segmentation annotations. There are 7 classes: \emph{window}, \emph{door}, \emph{balcony}, \emph{shop}, \emph{facade}, \emph{sky}, \emph{roof}. We split the whole dataset into train set, containing 113 images, and test set, containing 106 images , same as the paper \cite{2020PyramidALKNet} .

\textbf{Oxford RobotCar Facade.}\
The Oxford RobotCar Facade dataset contains a total of 500 deformed images. There are 5 classes: \emph{window}, \emph{door}, \emph{balcony}, \emph{shop}, \emph{facade}. We use the given benchmark data split, where 400 images are set as the train set and 100 images are set as the test set.

\subsection{Training Configuration}\label{subsec:Training Configuration}

We use Adam \cite{kingma2014adam} as the optimizer with learning rate 2e-4 and weight decay 1e-4. The data augmentation strategies we use include random color jittering, random horizontal flipping, and random scaling. We use the overall loss function defined in \cref{eq:finalloss} with $\alpha=1/9$ and $T=0.5$. We use the batch size of 4 and maximum input size of 1088$\times$1088 during training.

For the backbone, we select the ResNet-50 \cite{ResNet} that is already pretrained on the ImageNet dataset\cite{deng2009imagenet}, which is the same as the setting applied in \cite{2020PyramidALKNet} and \cite{2017DeepFacade}. Same as the setting in DeepLabV3\cite{deeplabv3}, we remove the strides of the last two stages (stage 4 and stage 5). Thus, the output stride of the last feature map is $8$, and this feature map is subsequently used for semantic segmentation using the vanilla DeepLabV3 segmentation classifier. 
As for the bbox detection, following Faster R-CNN\cite{FasterRCNN}, the feature maps from stage 2 and stage 5 followed by the Feature Pyramid Network\cite{fpn} are extracted for the generalized bbox detection.

We implement the whole network structure on the Pytorch \cite{paszke2019pytorch} platform and use one RTX A5000 GPU as the training hardware.

\begin{table}[]\footnotesize
\centering
\begin{tabular}{lcccc}
\toprule
\multirow{2}{*}{Model} & \multicolumn{2}{c}{RueMonge2014} &  \multicolumn{2}{c}{Oxford} \\
 & mIoU & Accuracy & mIoU & Accuracy \\
\midrule
Baseline & 70.37 & 87.5 & 51.8 &    94.41   \\
Mask R-CNN & 70.61 & 87.55 & 52.89 & 94.57 \\
Faster R-CNN & 70.1 & 87.2 & 52.25 &     94.36    \\
Facade R-CNN & \textbf{70.76} & \textbf{87.6}   &  \textbf{52.97}  &    \textbf{94.6}    \\
\bottomrule
\end{tabular}
\caption{Comparison of three R-CNNs under the threshold of 0.5 on the RueMonge2014 and the Oxford datasets, where all three models use the same fusion strategy described in \cref{subsec:mul learning}.}
\label{tab:Ablation study ge bbox on oxford}
\end{table}

\begin{center} 
\begin{figure}[htbp]
    \centering
    \includegraphics[width=0.45\textwidth]{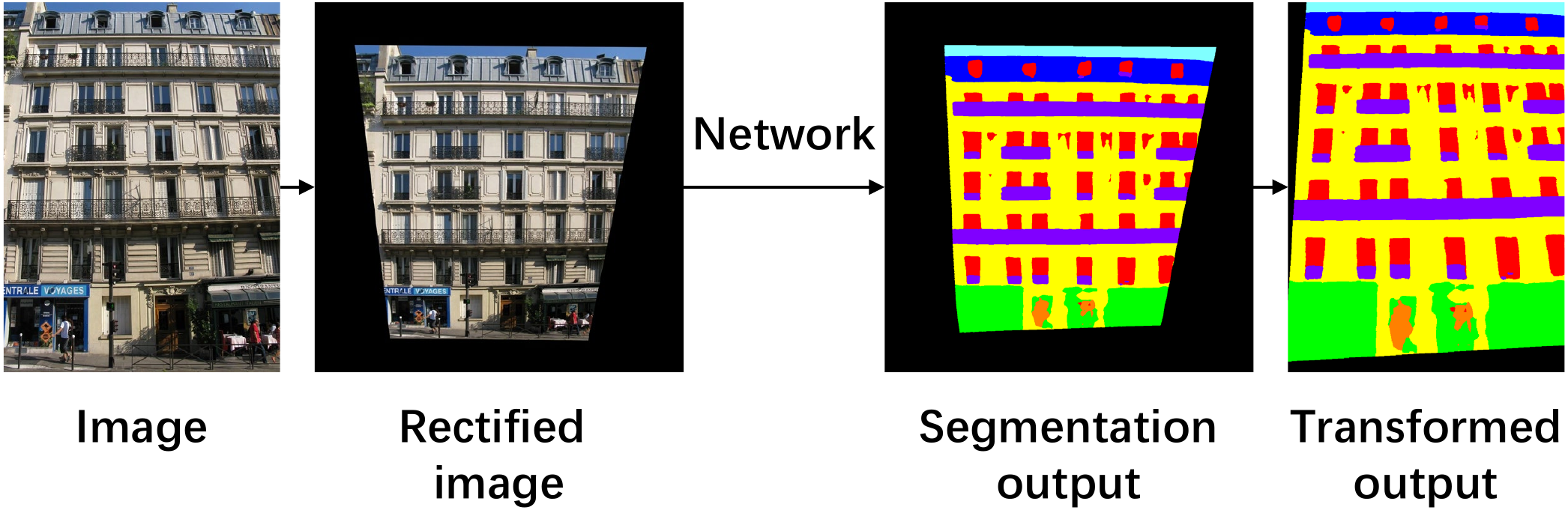}
    \caption{Illustration of rectification.} 
    \label{fig supp:rectification}
\end{figure}
\end{center}

\begin{table}[]
\centering
\begin{tabular}{lcc}
\toprule
Method & mIoU & Accuracy \\
\midrule    
Baseline & \textbf{70.37} & \textbf{87.5}\\
Rectification & 69.01 & 87.26 \\
\bottomrule
\end{tabular}
\caption{Performance of rectification.}
\label{tab:rectification}
\end{table}

\section{More Ablation Study}

\subsection{Generalized Bounding Box Detection}
We compare our Facade R-CNN with Mask R-CNN and Faster R-CNN in terms of fusion performance, where all three R-CNNs use the same baseline segmentation output for fair refinement comparison. From \cref{tab:Ablation study ge bbox on oxford}, we observe that Facade R-CNN outperforms the competitive counterpart Mask R-CNN by $+0.15$ mIoU and $+0.05$ accuracy on the RueMonge2014 dataset. As for the Oxford dataset, we obtain comparable performance as Mask R-CNN, while using less than 1/30 Flops and 1/150 memory consumption as illustrated in \cref{subsec:gbbox}.

\subsection{Rectification}

Except for directly applying parsing models on the deformed facade images, one would also first rectify the facade and then parse the undistorted one as shown in \cref{fig supp:rectification}. In this section, we test the performance when this pre-processing technique is introduced. In \cref{tab:rectification}, the rectification strategy is not able to bring improvement compared to the baseline.

\section{Visualization}\label{sec:visualization}

\begin{center} 
\begin{figure}[htbp]
    \centering
    \includegraphics[width=0.47\textwidth]{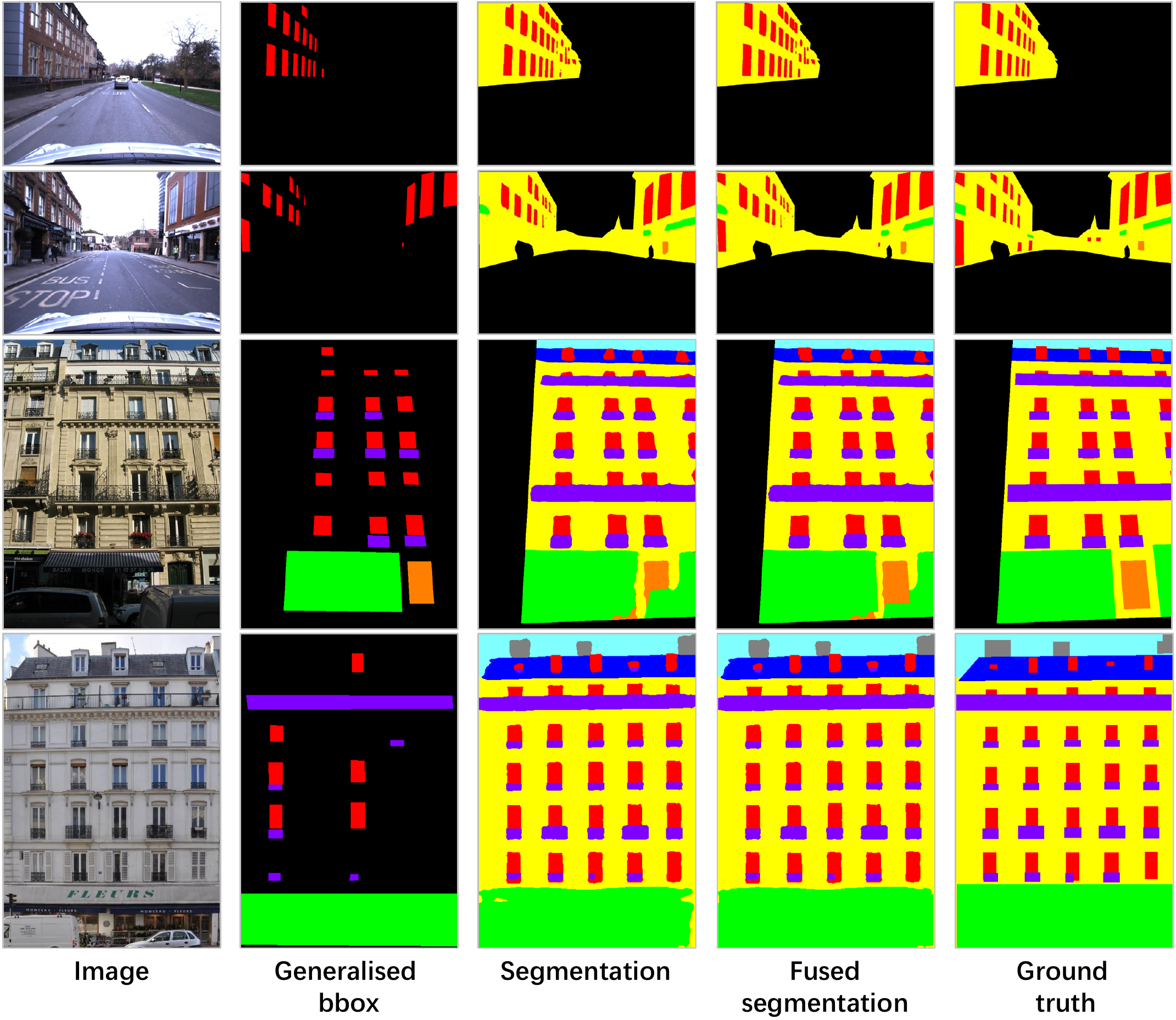}
    \caption{Parsing visualization.} 
    \label{fig supp:visualization}
\end{figure}
\end{center}

We report some of the parsing results in \cref{fig supp:visualization}, where the first two rows are for the Oxford RobotCar Facade dataset, the third row is for the RueMonge2014 dataset, and the last row is for the ECP dataset. As shown in \cref{fig supp:visualization}, our proposed generalized bbox is able to deal with the deformed facades and output the mask of the distorted objects, which could serve as a helpful module to refine the pure semantic segmentation output.

\bibliographystyle{named}
\bibliography{./bib/IEEEabrv,./bib/StringDefinitions,./bib/ref.bib}

\end{document}